\documentclass{article}

\usepackage{arxiv}

\usepackage[utf8]{inputenc} % allow utf-8 input
\usepackage[T1]{fontenc}    % use 8-bit T1 fonts
\usepackage{hyperref}       % hyperlinks
\usepackage{url}            % simple URL typesetting
\usepackage{booktabs}       % professional-quality tables
\usepackage{amsfonts}       % blackboard math symbols
\usepackage{nicefrac}       % compact symbols for 1/2, etc.
\usepackage{microtype}      % microtypography
\usepackage{lipsum}
\usepackage{graphicx}
\usepackage{algorithm2e}
\usepackage{cite}

\usepackage{lineno}
\usepackage[onehalfspacing]{setspace}

\usepackage{amsmath}

\title{Imitating by generating:  deep generative models for imitation of interactive tasks}

\author{
  Judith B\"utepage$^1$ \thanks{Corresponding author} \\
  \texttt{butepage@kth.se} \\
  \And
  Ali Ghadirzadeh $^{1,2}$ \\
  \texttt{algh@kth.se} \\
  \And
  \"Ozge \"Oztimur Karada\~g\,$^{1,3}$ \\
  \texttt{ozg@kth.se} \\
  \AND
  M\aa rten Bj\"orkman\,$^{1}$ \\
  \texttt{celle@kth.se} \\
  \And
  Danica Kragic\,$^{1}$ \\
  \texttt{dani@kth.se} \\ \\
  \AND
  \\
  $^1$ Division of Robotics, Perception and Learning \\
  KTH Royal Institute of Technology, Stockholm, Sweden \\ \\
  $^2$ Intelligent Robotics Research Group \\
  Aalto University, Espoo, Finland \\ \\
  $^3$ Department of Computer Engineering  \\
  Alanya Alaaddin Keykubat University, Antalya, Turkey
}

\begin{document}
\maketitle

\begin{abstract}

To coordinate actions with an interaction partner requires a constant exchange of sensorimotor signals. Humans acquire these skills in infancy and early childhood mostly by imitation learning and active engagement with a skilled partner. They require the ability to predict and adapt to one's partner during an interaction. In this work we want to explore these ideas in a human-robot interaction setting in which a robot is required to learn interactive tasks from a combination of observational and kinesthetic learning.  To this end, we propose a deep learning framework consisting of a number of components for (1) human and robot motion embedding, (2) motion prediction of the human partner and (3) generation of robot joint trajectories matching the human motion. To test these ideas, we collect human-human interaction data and human-robot interaction data of four interactive tasks "hand-shake", "hand-wave", "parachute fist-bump" and "rocket fist-bump".  We  demonstrate experimentally the importance of predictive and adaptive components as well as low-level abstractions to successfully learn to imitate human behavior in interactive social tasks. 

\end{abstract}

% keywords can be removed
\keywords{Imitation Learning and Human-robot Interaction}

% -----------------------------------------------------------------------------------
% Introduction
% -----------------------------------------------------------------------------------
\section{Introduction}
\label{sec:introduction}

% overall aim
Physical human-robot interaction requires the robot to actively engage in joint action with human partners. In this work, we are interested in robotic learning of physical human-robot tasks which require coordinated actions. We take inspiration from psychological and biological research and investigate how observational and kinesthetic learning can be combined to learn specific coordinated actions, namely interactive greeting gestures. 

% intro to joint action
In a more general context, coordinated actions between humans can be of functional nature, such as handing over an object, or of social importance, such as shaking hands as a greeting gesture. Thus, joint actions encompass any kind of coordination of actions in space and time in a social context. In general, joint actions require the ability to share representations, to predict others' actions and to integrate these predictions into action planning \cite{sebanz2006joint}. On a sensorimotor level coordinated actions require a constant coupling between the partners' sensory and motor channels \cite{vesper2017joint}. We aim at making use of sensorimotor patterns to enable a robot to engage with a human partner in actions that require a high degree of coordination such as hand-shaking.    

% imitation learning
The acquisition of the ability to engage in joint action during human infancy and early childhood is an active field of research in psychology \cite{brownell2011early}. Interaction is mostly learned in interaction, from observation, active participation or explicit teaching. While cultural differences exist, children are commonly presented with the opportunity to learn through guided participation in joint action with more experienced interaction partners \cite{rogoff1993guided}. In the robotics community two prominent techniques to learn actions from others are \textit{learning from demonstration} and \textit{imitation learning} \cite{billard2008robot, osa2018algorithmic}. Learning from demonstration can be seen as a form of imitation learning. It requires a teacher to intentionally demonstrate to a learner how an action should be performed. In a robotic learning scenario, this can imply direct kinesthetic teaching of trajectories. General imitation learning on the other hand includes also learners who passively observe an action and replicate it without supervision. When observing a human, who often has a different set of degrees of freedom, the robotic system first needs to acquire a mapping between embodiments before a motion can be imitated \cite{alissandrakis2007correspondence}.   

\begin{figure}[t]
    \centering
    \includegraphics[width=\textwidth]{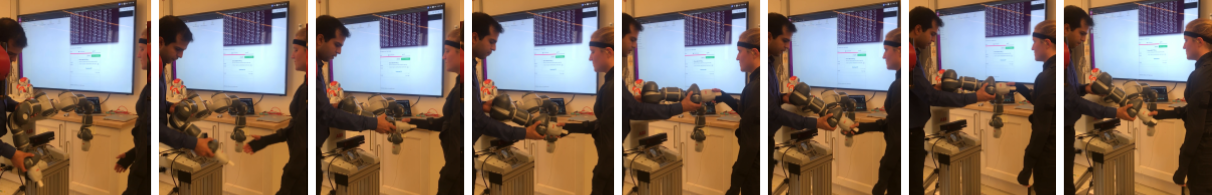}
    \caption{Kinesthetic teaching of a human-robot hand shake. The human partner is wearing a motion capture suit to record joint positions.}
    \label{fig:hand_shake_teaching}
\end{figure}

% short summary
In this work, we are interested in teaching a robot to coordinate with a human in time and space. Therefore we require adaptive and predictive models of sensorimotor patterns such as joint trajectories and motor commands of interactive tasks. To this end, we develop deep generative models that represent joint distributions over all relevant variables over time. The temporal latent variables in these models encode the underlying dynamics of the task and allow for a sensorimortor coupling of the human and the robot partner. As depicted in Figure \ref{fig:hand_shake_teaching}, collecting data by kinesthetic teaching for human-robot interaction tasks is tedious and time-consuming. We propose to first model the dynamics of human-human interaction and subsequently use the learned representation to guide the robot's action selection during human-robot interaction. 

% next up
Before diving into the theory, in the next section we will shortly introduce the field of robotic imitation learning and point out how the general field differs from the requirements needed for imitation learning for interaction. Finally, we will motivate our choice of model and explain the basic assumptions of deep generative models. 

% n learning and point ouature of tasks - from objectsation to joint action, learning from demonstration, learning from observation, learning joint action from observation

%Always cite with $cite$. No subsections in the intro.

\section{Background}
\label{sec:background}

Traditionally, robotic imitation learning is applied to individual tasks in which the robot has to acquire e.g. motor skills and models of the environment. Our goal is to extend the ideas to interactive settings in which a human partner has to be incorporated into action selection.  Thus. we aim at transferring knowledge form observing human-human interaction (HHI) to human-robot interaction (HRI).

\subsection{Robotic imitation learning of trajectories}
\label{sec:robotimitationlearning}

Imitation learning is concerned with acquiring a policy, i.e. a function that generates the optimal action given an observed state. While reinforcement learning usually solves this task with help of active exploration by the learning agent, in imitation learning the agent is provided with observations of states and actions from which to learn. These demonstrations can either be generated in the agent's own state space, e.g. by tele-operation \cite{argall2009survey}, or in the demonstrators embodiment, e.g. a human demonstrating actions for a robot. In this work we combine these approaches to teach a robot arm trajectories required for a number of interactive tasks.

Learning trajectory generating policies from demonstration has been addressed with for example a combination of Gaussian Mixture Models and Hidden Markov Models \cite{calinon2010learning}, probabilistic flow tubes \cite{Dong2011MotionLI,Dong2012LearningAR} or probabilistic motion primitives \cite{maeda2017probabilistic}. The general strategy in this case is to first gather training data in form of trajectories and to align these temporally e.g. with the help of Dynamic Time Warping \cite{Sakoe1978DynamicPA}. Once the training data has been pre-processed in this way, the model of choice is trained to predict the trajectory of robotic motion for a certain task. During employment of the model, the online trajectory needs to  be aligned with the temporal dynamics of the training samples in order to generate accurate movements. Depending on the trajectory representation, e.g. torque commands or Cartesian coordinates, the model predictions might be highly dependent on the training data. For example, when the task is to learn to grasp an object at a certain location, the model might not generalize to grasping the same object at a different location. 

This constant need of alignment and reliance on demonstrations hampers the models to work in a dynamic environment with varying task demands. For example, if the task is to shake hands with a human, the number of shaking cycles and the length of each individual shake can vary from trial to trial and have to be estimates online instead of being predicted once prior to motion onset. These requirements for online interaction are discussed in more detail below.

\subsection{Requirements for online interaction}
\label{sec:requirementts}

Interaction with humans requires a robotic system to be flexible and adaptive \cite{dautenhahn2007socially, maeda2017phase}. To meet these requirements, the robot needs to be able to anticipate future human actions and movements \cite{koppula2015anticipating}. Thus, imitation learning for interaction is different from non-social imitation learning as it requires to learn a function not only of one's own behavior, but also of the partner's behavior. 

These requirements stand therefore in contrast to the approaches to imitation learning discussed in Section \ref{sec:robotimitationlearning} which focus on learning a trajectory of a fixed size. \cite{maeda2017phase} address the problem of adjusting to the speed of the human's actions by introducing an additional phase variable. This variable can be interpreted as an indication of the progress of the movement of the human to which the robot has to adapt. However, such an approach is only feasible for interactions which require little mutual adaptation beyond speed. For example, during  a hand-shake interaction, it is not only important to meet the partner's hand at an appropriate time, but also to adjust to the frequency and height of every up-and-down movement. Thus, online interaction requires the prediction of the partner's future movements in order to adapt to them quickly and a constant update of these predictions based on sensory feedback. 

An additional requirement for natural human-robot interaction is to provide fluent coordination. We envision robot's to actively engage in an interaction such that the human partner does not have to wait with a stretched arm until the robot reacts and moves its arm to engage in a hand-shake. Making use of predictive models allows the robot to initiate its movements before the human has reached the goal location. These models also provide a basis for collision-free path planning to assure safe interaction in shared workspaces.

Since humans are involved in the data collection process and kinesthetic teaching is time consuming and requires expert knowledge, the amount of training data is restricted. Therefore, any method used to learn trajectories must be data efficient. Many modern imitation learning techniques build on ideas from deep reinforcement learning \cite{zhang2018deep, li2017infogail} which is often data intensive. We rely on a  model class which is regularized by its Bayesian foundation and therefore less prone to overfit to small datasets. This model class of deep latent variable models has been mostly used to model images. Here, we take inspiration from earlier work  in which we model human motion trajectories \cite{butepage2018anticipating} and robot actions \cite{ghadirzadeh2017deep} with help of deep generative models. We extend the ideas to represent the dynamics of human-robot interaction in a joint model. For those unfamiliar with the ideas of Variational Autoencoders, we introduce the underlying concept of this model class below.

\subsection{Deep generative models and inference networks}
\label{sec:dgm}

In this work, we model human and robotic motion trajectories with help of Variational Autoencoders (VAEs) \cite{kingma2013auto, rezende2014stochastic}, a class of deep generative models. In contrast to Generative Adversarial Networks \cite{Goodfellow2014GenerativeAN} and flow-based methods \cite{dinh2016density, kingma2018glow}, VAEs allow us to define our assumption in terms of a probabilistic, latent variable model in a principled manner.  While we focus on the main concepts and the mathematical foundations of VAEs,  we refer the reader to \cite{zhang2018advances} for an in-depth review on modern advances in variational inference and VAEs. In the next section, we will shortly introduce the concepts of variational inference.  

\subsubsection{Variational inference}

To begin with, we assume that the observed variable, or data point, $\mathbf{x} \in \mathbb{R}^{d_{x}}$ depends on latent variables $\mathbf{z} \in \mathbb{R}^{d_{z}}$. If the dataset consists of images, the latent variables or factors $\mathbf{z}$ describe the objects, colors and decomposition of the image. If, as we will introduce later, the dataset consists of human or robot joint movements, the underlying factors describe the general movement patterns. For example, a waving movement, in which many joints are involved, can be described by a single oscillatory latent variable. Usually the dimension of $\mathbf{z}$ is smaller than the dimension of $\mathbf{x}$, i.e. $d_z<<d_x$. 

In general, this model describes a joint distribution over both variables $p_\theta(\mathbf{x}, \mathbf{z}) = p_\theta(\mathbf{x}| \mathbf{z})p_\theta(\mathbf{z})$ where $\theta$ are the parameters.  This modelling assumption allows us to generate new observations with help of the mathematical model instead of employing a physical system. First, a latent variable is sampled, from a prior distribution $\mathbf{z} \sim p_\theta(\mathbf{z})$. For example, to generate a waving arm movement, we sample where in the oscillation the arm starts and the initial velocity. Then we sample the actual poses conditioned on these latent variables. The conditional distribution $\mathbf{x} \sim p_\theta(\mathbf{x}|\mathbf{z})$ encodes the mapping from the latent space to the observed space. Thus, the generative process looks as follows:
\begin{equation}
\mathbf{x} \sim p_\theta(\mathbf{x}|\mathbf{z}), \ \mathbf{z} \sim p_\theta(\mathbf{z}).
\end{equation}
In order to determine the structure of the latent variables that generated on observed set  consisting of $n$ data points $\mathbf{X} = \{\mathbf{x}_i\}_{i=1:n}$, one requires access to the posterior distribution $p_\theta(\mathbf{z}_i|\mathbf{x}_i)$ for each data point $\mathbf{x}_i$. Often exact inference of this term is intractable which is why one recedes to approximate inference techniques such as Monte Carlo sampling and variational inference (VI).  VAEs combine VI for probabilistic models with the representational power of deep neural networks. VI is an optimization based inference technique which estimates the true posterior distribution $p_\theta(\mathbf{Z}|\mathbf{X})$ with a simpler approximate distribution $q_\phi(\mathbf{Z})$ where $\phi$ are parameters and $\mathbf{Z} = \{\mathbf{z}_i\}_{i=1:n}$ is the set of latent variables corresponding to the data set. A common approach is the mean-field approximation which assumes that the latent variables are independent of each other $q_\phi(\mathbf{Z}) = \prod_{i=1}^n q_\phi(\mathbf{z}_i)$. As an example, if $q_\phi(\mathbf{z}_i)$ follows a Gaussian distribution, we need to identify a  mean $\mu_i$ and variance $\sigma_i$ for  every data point $q_\phi(\mathbf{Z}) = \prod_{i=1}^n \mathcal{N}(\mu_i, \sigma_i)$. For the entire dataset $(\mathbf{X}, \mathbf{Z})$, the parameters of this distribution are determined by optimizing the Evidence Lower BOund (ELBO)

\begin{equation}
\log p_\theta(\mathbf{X}) \geq \mathbb{E}_{q_\phi(\mathbf{Z})} \log \frac{p_\theta(\mathbf{X},\mathbf{Z})}{q_\phi(\mathbf{Z})} = \mathbb{E}_{q_\phi(\mathbf{Z})} \log  p_\theta(\mathbf{X}|\mathbf{Z}) - D_{KL}(q_\phi(\mathbf{Z})||p_\theta(\mathbf{Z})),  \label{eq:VI_elbo}
\end{equation} 
where $D_{KL}(q||p) = \mathbb{E}_q \log \frac{q}{p}$ is a distance measure between two distributions $q$ and $p$. 

Traditional VI approximates a latent variable distribution $q_\phi(\mathbf{z}_i)$ for every data point $i$ which becomes expensive or impossible when the number of data points $n$ is large. VAEs circumvent this problem by learning a direct functional mapping from the data space to the latent space and vice versa. We will detail this method in the next section.

\subsubsection{Variational Autoencoders}

Instead of approximating $n$ sets of parameters, VAEs employ so called inference networks to learn a functional mapping from the data space into the latent space. Thus we define each latent variable to be determined by a distribution  $\mathbf{z}_i \sim q_\phi(\mathbf{z}_i|\mathbf{x}_i)$ which is parameterized by a neural network (the inference network) that is a function of the data point $\mathbf{x}_i$. In the Gaussian case this would imply that $\mathbf{z}_i \sim \mathcal{N}(\mu(\mathbf{x}_i), \sigma(\mathbf{x}_i))$, where $\mu(\cdot)$ and $\sigma(\cdot)$ are neural networks mapping from the data space to the parameter space of the latent variables. Likewise, the likelihood is represented by neural network mappings (the generative network) $\mathbf{x}_i \sim p_\theta(\mathbf{x}_i|\mathbf{z}_i)$. In this way, VAEs do not estimate $n$ sets of parameters but only the parameters of the inference and generative network. These are optimized with help of the ELBO 
\begin{equation}
\log p_\theta(\mathbf{X}) \geq  \mathcal{L}(\mathbf{X}, \theta, \phi) = \frac{1}{n} \sum_{i=1}^n \mathbb{E}_{q_\phi(\mathbf{z}_i|\mathbf{x}_i)} \log  p_\theta(\mathbf{x}_i|\mathbf{z}_i) - D_{KL}(q_\phi(\mathbf{z}_i|\mathbf{x}_i)||p_\theta(\mathbf{z}_i)). \label{eq:VAE_elbo}
\end{equation} 
Note that we replaced the expectation in Equation \ref{eq:VI_elbo} with the Monte Carlo estimate summing over the individual data points.

\section{Methodology}
\label{sec:material_and_methods}

%\begin{center}
%\begin{tabular}{@{}ll@{}} 
%\textbf{Symbol} & \textbf{Description} \\
%$\mathbf{x}^{a}_{t:t'}$ & $t'-t$ frames of $d_x$ joint positions of agent $a$ \\
%$\mathbf{z}^a_{t}$ &   $d_z$ dimensional latent variable of agent $a$ \\
%$\theta = (\theta_x,\theta_z,\theta_s)$  & neural network parameters
%\end{tabular}
%\end{center}
 
Following the introduction of VAEs above, we will now detail how we employ them to learn the sensorimotor patterns required for interactive tasks. We will begin with a description of human-human dynamics modelling which is subsequently used to guide the human-robot interaction model.

% --------------------------------------------------------------------------------------------------------
% A generative model of interaction
% --------------------------------------------------------------------------------------------------------
\subsection{A generative model of interaction}
\label{sec:generative_model}

\begin{figure}[t]
    \centering
    \includegraphics[ width=0.7\textwidth]{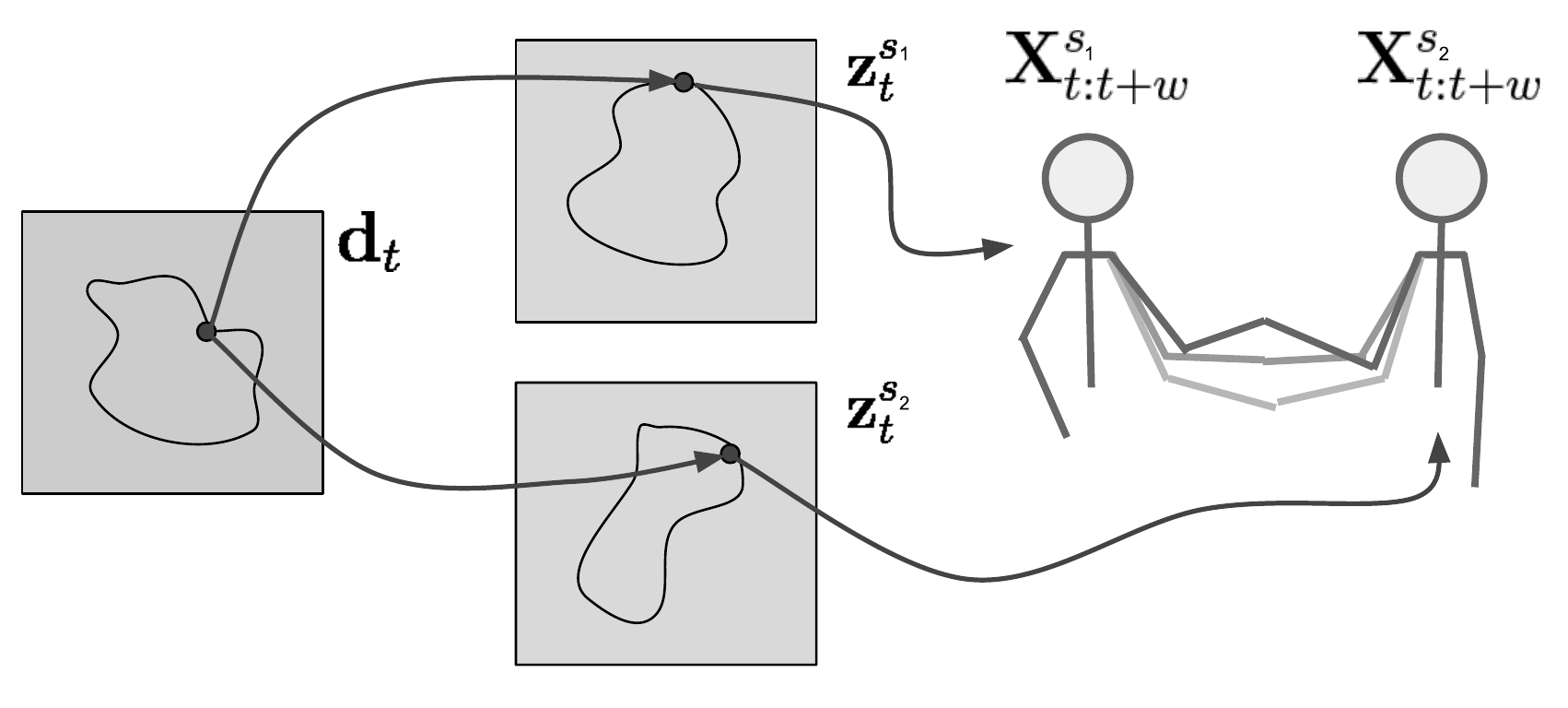}
    \caption{The task dynamics $\mathbf{d}_{t}$ govern the activity of the latent variables of both partners $z^{s_1}_t$ and $z^{s_2}_t$. These in turn determine the future movement of the partners $\mathbf{x}^{s_1}_{t:t+w}$  and $\mathbf{x}^{s_2}_{t:t+w}$.}
    \label{fig:dynamics}
\end{figure}

In general we assume that a recording $rec$ consists of $T_{rec}$ observations $\mathbf{x}^{s_1}_{1:T_{rec}}$ and $\mathbf{x}^{s_2}_{1:T_{rec}}$, where $(s_1,s_2) = (human_1, human_2)$, and $\mathbf{x}^{s}_{t}$ represents a single frame containing the joint positions of human $s \in \{s_1,s_2\}$. During testing time, we would like to be able to infer a future window (of size $w$) of observations after time $t$, i.e. we would like to predict $\mathbf{x}^{s_1}_{t:t+w}$ and $\mathbf{x}^{s_2}_{t:t+w}$. 
We assume a generative process that looks as follows
\begin{align}
\mathbf{x}^{s_1}_{t:t+w} \sim & \   p_{\theta_x}(\mathbf{x}^{s_1}_{t:t+w}|\mathbf{z}^{s_1}_{t}), \quad \mathbf{z}^{s_1}_{t} \sim p_{\theta_z}(\mathbf{z}^{s_1}_{t}|\mathbf{d}_{t}), \quad \mathbf{d}_{t} \sim p_{\theta_s}(\mathbf{d}_{t}|h^{s_1}_t), \quad h^{s_1}_t = f_\psi(h^{s_1}_{t-1}, \mathbf{x}^{s_1}_{t-1}) \\ 
\mathbf{x}^{s_2}_{t:t+w} \sim &  \   p_{\theta_x}(\mathbf{x}^{s_2}_{t:t+w}|\mathbf{z}^{s_2}_{t}), \quad \mathbf{z}^{s_2}_{t} \sim p_{\theta_z}(\mathbf{z}^{s_2}_{t}|\mathbf{d}_{t}), \ \quad \mathbf{d}_{t} \sim p_{\theta_s}(\mathbf{d}_{t}|h^{s_2}_t), \quad h^{s_2}_t = f_\psi(h^{s_2}_{t-1}, \mathbf{x}^{s_2}_{t-1}). \label{eq:generative_model}
\end{align}
Here, the latent variables $z^{s_1}_t$ and $z^{s_2}_t$ for agent $s_1$ and $s_2$ encode the next time window  $\mathbf{x}^{s_1}_{t:t+w}$ and $\mathbf{x}^{s_2}_{t:t+w}$, while $h^{s_2}_t$ is the deterministic output of a recurrent model $f$. As depicted in Figure \ref{fig:dynamics}. $\mathbf{d}_{t}$ encodes shared task dynamics which can be derived from the movement of either subject independently. These shared dynamics influence how each partner moves through $z^{s_1}_t$ and $z^{s_2}_t$. In summary, the generative model for agent $s_1$ represents the joint distribution $p_\theta(\mathbf{x}^{s_1}_{t:t+w}, \mathbf{z}^{s_1}_{t}, \mathbf{d}_{t}|h^{s_1}_t)$ conditioned on a deterministic summary of the past $h^{s_1}_t$ and parameterized by $\theta = (\theta_x,\theta_z,\theta_s)$.

In the following, we will describe how to learn each of the components for human-human interaction and subsequently how to transfer this knowledge to a human-robot interaction scenario.

\begin{figure}[t]
    \centering
    \includegraphics[trim={0.1cm 0.1cm 0 0},clip,width=0.8\textwidth]{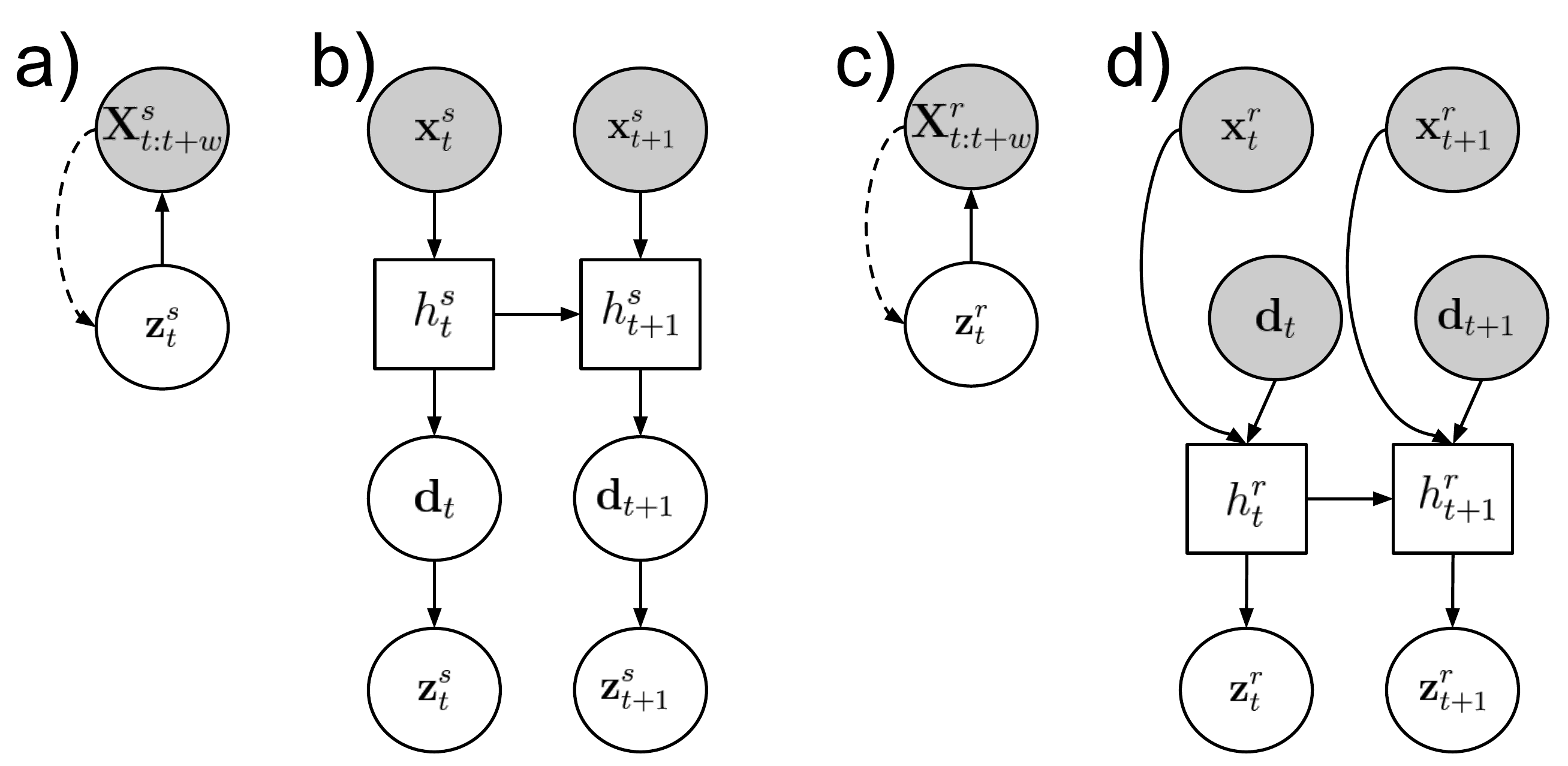}
    \caption{An overview of the model structure. \textbf{a)} Human motion embedding,  \textbf{b)} Task dynamics model,  \textbf{c)} Robot motion embedding,  \textbf{d)} Human motion embedding,  Interaction model with predictive input. Gray circles represent observed variables, white circles are unobserved variables and a white square indicates a deterministic unit. A filled line shows the generative process while dotted lines indicate inference connections.}
    \label{fig:models}
\end{figure}

% --------------------------------------------------------------------------------------------------------
% Encoding motion dynamics
% --------------------------------------------------------------------------------------------------------
\subsubsection{Motion embeddings}
\label{sec:encoding_motion_dynamics}

One problem when it comes to predicting the future is that there exist many possible ones. When using a mean-squared error based cost function during training, this will lead the model to rely on predicting only the average, not many different trajectories.  
We will circumvent this problem by first learning a latent space that encodes the future without knowledge of the past and then to learn a distribution over the latent variables which is conditioned on the past (e.g. $p(\mathbf{z}^{s_1}_{t}|\mathbf{d}_{t})$ in Equation \ref{eq:generative_model}). 
At each time step, we assume that there exists latent variables $z^{s_1}_t$ and $z^{s_2}_t$ for agent $s_1$ and $s_2$ which encode the next time window  $\mathbf{x}^{s_1}_{t:t+w}$ and $\mathbf{x}^{s_2}_{t:t+w}$. We assume that both humans are encoded into a common space, therefore we will replace the superscripts $s_1$ and $s_2$ with $s$ in the following discussion. 

To infer the latent variables efficiently from data, we apply variational autoencoders (introduced in Section \ref{sec:dgm}). 
To this end, we define the following generative process:

\begin{align}
\mathbf{x}^s_{t:t+w} \sim & \   p_{\theta_x}(\mathbf{x}^s_{t:t+w}|\mathbf{z}^s_{t}), \quad \mathbf{z}^s_{t} \sim p_{\theta_z}(\mathbf{z}^s_{t}) = \mathcal{N}(0,1), \text{ and approximate posterior }  \mathbf{z}^s_{t}  \sim  q_{\phi_z}(\mathbf{z}^s_{t}|\mathbf{x}^s_{t:t+w}).
\end{align}

The graphical model is depicted in Figure \ref{fig:models} a).
The parameters $(\theta_x,\phi_z)$ of the generative network $p_\theta(\mathbf{x}^s_{t:t+w}|\mathbf{z}^s_{t})$ and the inference network $q_{\phi_z}(\mathbf{z}^s_{t}|\mathbf{x}^s_{t:t+w})$ are jointly trained on the training data collected from both humans to optimize the Evidence Lower BOund (ELBO) 

\begin{align}
\mathcal{L}(\mathbf{x}^s_{t:t+w}, \theta, \phi) = \mathbb{E}_{q_{\phi_z}(\mathbf{z}^s_{t}|\mathbf{x}^s_{t:t+w})} \log p_{\theta_x}(\mathbf{x}^s_{t:t+w}|\mathbf{z}^s_{t}) - D_{KL}(q_{\phi_z}(\mathbf{z}^s_{t}|\mathbf{x}^s_{t:t+w})||p_{\theta_x}(\mathbf{z}^s_{t})). \label{eq:human_vae_loss}
\end{align}

% --------------------------------------------------------------------------------------------------------
% Encoding task dynamics
% --------------------------------------------------------------------------------------------------------
\subsubsection{Encoding task dynamics}
\label{sec:encoding_task_dynamics}

Once the motion embeddings have been learned, the whole generative model in Equation \ref{eq:generative_model}, as depicted in Figure \ref{fig:models} b), can be trained. To this end, we need to infer the parameters $(\theta_z, \theta_s, \psi)$ to estimate $p_{\theta_z}(\mathbf{z}^{s}_{t}|\mathbf{d}_{t}), p_{\theta_s}(\mathbf{d}_{t}|h^{s}_t)$ and $f_\psi(h^{s}_{t-1}, \mathbf{x}^{s}_{t-1})$. 

The loss function is defined as follows
\begin{align}
\mathcal{S}(\mathbf{x}^{s_1}_{t-1:t+w}, \mathbf{x}^{s_2}_{t-1:t+w}, \theta_z, \theta_s, \psi) =  & \sum_{s \in \{s_1, s_2 \}} D_{KL}(p_{\theta_z}(\mathbf{z}^{s}_{t}|\mathbf{d}_{t}) ||q_{\phi_z}(\mathbf{z}^{s}_{t}|\mathbf{x}^{s}_{t:t+w}))  + \nonumber \\
&JSD(p_{\theta_s}(\mathbf{d}_{t}|h^{s_1}_t)||p_{\theta_s}(\mathbf{d}_{t}|h^{s_2}_t)). \label{eq:task_loss}
\end{align}

The first line in Equation \ref{eq:task_loss} forces the distributions over latent variables $\mathbf{z}^s_t$ that depend on the past to be close to the expected motion embedding at time $t$.
The second line enforces that the latent variable $\mathbf{d}_t$, which encodes the task dynamics are the same for both agents. As the KL divergence is not symmetric, we use here the Jensen–Shannon divergence, which is defined as $JSD(p||q) = \frac{1}{2}(D_{KL}(p||\frac{1}{2}(p+q)) + D_{KL}(q||\frac{1}{2}(p+q)))$  for two distributions $p$ and $q$.

\subsubsection{Interactive embodiment mapping}
\label{sec:Mapping between embodiments}

Once trained, the generative model described above can be used to generate future trajectories for both agents given that only one agent has been observed. This is achieved by e.g. predicting the task dynamics variable $\mathbf{d}_{t} \sim p_{\theta_s}(\mathbf{d}_{t}|h^{s_1}_t)$ with help of data collected for agent $s_1$ and using this variable to infer both $\mathbf{z}^{s_1}_{t} \sim p_{\theta_z}(\mathbf{z}^{s_1}_{t}|\mathbf{d}_{t})$ and $\mathbf{z}^{s_2}_{t} \sim p_{\theta_z}(\mathbf{z}^{s_2}_{t}|\mathbf{d}_{t})$. We will make use of this fact to infer not only a human partner's future movement, but also to guide how a robotic partner should react given the observed human. 

As training data acquisition with a robot and a human in the loop is cumbersome and time consuming, we do not have access to as much training data of the human-robot interaction compared to the human-human interaction. Therefore, we will leverage the task dynamics representation learned from human-human interaction to guide the robot's corresponding motion commands. To this end, we extract the task dynamics distribution from the human partner for each time step of the human-robot interaction recordings and learn a mapping to the robot's motion commands with a second dynamics model.

In more detail, given a recording $r$ which consists of $T_r$ observations $\mathbf{x}^{s_1}_{1:T_r}$ and $\mathbf{x}^r_{1:T_r}$, where $\mathbf{x}^r_{t}$ represents the robot's state at time $t$, we first collect $\mathbf{d}_{1:T_r}$ which we set to the mean of the distribution $p_{\theta_s}(\mathbf{d}_{t}|h^{s_1}_t)$ for each time step $t$. We are now equipped with a training data set, containing the data point pairs $(\mathbf{x}^r_{t:t+w}, \mathbf{d}_{t})$. In order to learn a predictive model from the task dynamic variable $\mathbf{d}_{t}$ to the future motion commands of the robot, $\mathbf{x}^r_{t:t+w}$, we design a similar approach to the model described for human-human interaction. It includes a Variational Autoencoder functioning as a motion embedding and a recurrent network that encodes the robot motion over time. These two models are depicted in Figure \ref{fig:models} c) and d) respectively.

\noindent \textbf{Interaction model with predictive input:}
Similar to the human-human setting in Equation \ref{eq:generative_model}, the generative model for the robot motion is as follows
\begin{align}
\mathbf{x}^{r}_{t:t+w} \sim & \   p_{\theta_{xr}}(\mathbf{x}^{r}_{t:t+w}|\mathbf{z}^{r}_{t}), \quad \mathbf{z}^{r}_{t} \sim p_{\theta_{zr}}(\mathbf{z}^{r}_{t}|h^{r}_t), \quad h^{r}_t = f_{\psi_r}(h^{r}_{t-1}, \mathbf{x}^{r}_{t-1}, \mathbf{d}_{t-1})  \label{eq:generative_model_robot}
\end{align}

Just as in the human-human setting, we first train a motion embedding VAE on the robot data, i.e. we train the following model with the same loss function as in Equation \ref{eq:human_vae_loss}
\begin{align}
\mathbf{x}^r_{t:t+w} \sim & \   p_{\theta_{xr}}(\mathbf{x}^r_{t:t+w}|\mathbf{z}^r_{t}), \quad \mathbf{z}^r_{t} \sim p_{\theta_{zr}}(\mathbf{z}^r_{t}) = \mathcal{N}(0,1), \text{ and approximate posterior }  \mathbf{z}^r_{t}  \sim  q_{\phi_{zr}}(\mathbf{z}^r_{t}|\mathbf{x}^r_{t:t+w}). \label{eq:robot_vae_loss}
\end{align}

Subsequently, we assume that the parameters $(\theta_{zr},   \psi_r)$ in Equation \ref{eq:generative_model_robot} are inferred by optimizing
\begin{align}
\mathcal{S}(\mathbf{x}^{r_1}_{t-1:t+w}, \mathbf{d}_t, \theta_z,  \psi) =  &  D_{KL}(p_{\theta_{zr}}(\mathbf{z}^{s}_{t}|h^{r}_t) ||q_{\phi_{zr}}(\mathbf{z}^{s}_{t}|\mathbf{x}^{r}_{t:t+w})), \label{eq:task_loss_robot}
\end{align}
where the dynamics $\mathbf{d}_t \sim p_{\theta_s}(\mathbf{d}_{t}|h^{s_1}_t)$ are extracted with help of the models trained on the human-human data. We summarize the training procedure of all our model in Algorithm \ref{algo}.

\begin{algorithm}[t]
 \textbf{Human-human interaction} \\
 \hspace{0.5cm} \KwData{$\mathbf{X}^{s_1,s_2} = \{\mathbf{x}^{s_1}_{1:T_{rec}}, \mathbf{x}^{s_2}_{1:T_{rec}}\}_{rec \in \text{HHI recordings}}$} 
 \hspace{0.5cm} \textbf{Step 1: Human motion embedding} \\
  \hspace{1cm} Fit $p_{\theta_x}(\mathbf{x}^s_{t:t+w}|\mathbf{z}^s_{t})$ and $q_{\phi_z}(\mathbf{z}^s_{t}|\mathbf{x}^s_{t:t+w})$ to $\mathbf{X}^{s_1,s_2}$, following Equation \ref{eq:human_vae_loss}. \\
  \hspace{0.5cm} \textbf{Step 2: Task dynamics} \\
  \hspace{1cm}  Fit $p_{\theta_z}(\mathbf{z}^{s}_{t}|\mathbf{d}_{t}), p_{\theta_s}(\mathbf{d}_{t}|h^{s}_t)$ and $f_\psi(h^{s}_{t-1}, \mathbf{x}^{s}_{t-1})$ to $\mathbf{X}^{s_1,s_2}$, following Equation \ref{eq:task_loss}.  \\
  \textbf{Human-robot interaction} \\
   \hspace{0.5cm} \KwData{$\mathbf{X}^{s_1,r} = \{\mathbf{x}^{s_1}_{1:T_{rec}}, \mathbf{d}_{1:T_{rec}}, \mathbf{x}^{r}_{1:T_{rec}}\}_{rec \in \text{HRI recordings}}$ , where $\mathbf{d}_{t} = \text{ mean  of }  p_{\theta_s}(\mathbf{d}_{t}|h^{s}_t)$ }
  \hspace{0.5cm} \textbf{Step 3: Robot motion embedding} \\
  \hspace{1cm} Fit $p_{\theta_{xr}}(\mathbf{x}^r_{t:t+w}|\mathbf{z}^r_{t})$ and $q_{\phi_{zr}}(\mathbf{z}^r_{t}|\mathbf{x}^r_{t:t+w})$ to $\mathbf{X}^{s_1,r}$, combining Equation \ref{eq:human_vae_loss} and \ref{eq:robot_vae_loss}.  \\
  \hspace{0.5cm} \textbf{Step 4: Interactive embodiment mapping} \\
  \hspace{1cm}  Fit $p_{\theta_{zr}}(\mathbf{z}^{r}_{t}|h^{r}_t)$ and $f_{\psi_r}(h^{r}_{t-1}, \mathbf{x}^{r}_{t-1}, \mathbf{d}_{t-1}) $ to $\mathbf{X}^{s_1,r}$, following Equation \ref{eq:task_loss_robot}. \\
  \vspace{0.3cm}
 \caption{   All four steps of our combined motion embedding and dynamics modelling framework.} \label{algo}
\end{algorithm}

\subsection{Generating interactions}
\label{sec:motion_generation}

In order to employ our models during an ongoing interaction, we need to predict future time steps. As the dynamics and the motion embeddings encode a window of the next $w$ time steps, the prediction up to this horizon is straight forward as it only requires a propagation of the observed data. To go beyond a time frame of $w$ is made possible by our generative design. Instead of propagating observed data, one can let the models predict the next $w$ time frames based on the observed data and provide these as an input to the model. In case of the human-robot interaction model, one has to first predict the human's future motion to extract the matching dynamics variables and can subsequently use these variables together with predictions of the robot's motion to generate long-term robot motion. During online interaction these predictions can be updated on the fly when new data is observed. 

\subsection{Baselines}

We benchmark our approach on three baselines. Our own approach will be called \textit{Human Motion Embedding} in the following.

The first baseline tests whether our predictive and adaptive approach is necessary or whether more static imitation learning techniques suffice. To test this, we group the robot trajectories in the training data according to action type and use Dynamic Time Warping (DTW) to align them. We fit Gaussian distributions with full covariance matrices to the trajectory of each of the robot's joints. If DTW resulted in a trajectory length of $T_{DTW}$ for a certain action type and joint, then the Gaussian is of dimension $T_{DTW}$. A sample from each Gaussian model constitutes therefore a trajectory in joint angle space without input from  the current human movement. We call this approach \textit{Gaussian model}.

The second baseline tests whether our approach actually benefits from the encoded dynamics learned with the HHI data. Thus, in this case we train the same model as described in Section \ref{sec:Mapping between embodiments}. However, instead of feeding the dynamics variable $\mathbf{d}_t$  into the recurrent network $h^{r}_t = f_{\psi_r}(h^{r}_{t-1}, \mathbf{x}^{r}_{t-1}, \mathbf{d}_{t-1})$ in Equation \ref{eq:generative_model_robot}, we feed the current human joint position $\mathbf{x}^{s}_{t-1}$, i.e. $h^{r}_t = f_{\psi_r}(h^{r}_{t-1}, \mathbf{x}^{r}_{t-1}, \mathbf{x}^{s}_{t-1})$. This also affects the loss in Equation \ref{eq:task_loss_robot}, which now is a function of $\mathbf{x}^{s}_{t-1}$, i.e., $\mathcal{S}(\mathbf{x}^{r_1}_{t-1:t+w}, \mathbf{x}^{s}_{t-1}, \theta_{zr}, \psi_r)$. We call this approach \textit{Raw Data HR} which symbolizes that we provide raw human and robot data as input to the model.

The third baseline tests whether the human observation is required at all or whether the approach is powerful enough to predict based on robot joint position alone. In this case we train the same model as described in Section \ref{sec:Mapping between embodiments}, but provide only the current robot joint positions $\mathbf{x}^{r}_{t-1}$, i.e. $h^{r}_t = f_{\psi_r}(h^{r}_{t-1}, \mathbf{x}^{r}_{t-1})$. This also affects the loss in Equation \ref{eq:task_loss_robot}, i.e., $\mathcal{S}(\mathbf{x}^{r_1}_{t-1:t+w}, \theta_{zr}, \psi_r)$.
 We call this approach \textit{Raw Data R} which symbolizes that we provide only raw robot data as input to the model.

\section{Experimental setup and model fitting}

In this section we detail the experimental setup and model performances. For details regarding model architectures and model training, such as train and test splits, please see the Supplementary Material.

\subsection{Task description}
\label{sec:task_description}

Our interactive tasks consist of performing four different greeting gestures with a human partner. In each task execution we assume the identity of the gesture to be known apriori as the focus of this work lies on learning continuous interactive trajectories. However, our method can easily be extended to automatically infer the action type \cite{butepage2018classify}. 
Two of the gestures fall into the category of dyadic leader–follower interaction, while the other two partners carry equal roles.
The interactive gestures are the following: 
\begin{table}[h]
\centering
{\renewcommand\arraystretch{1.25}
\begin{tabular}{l|l} 
\multicolumn{2}{l}{\textbf{Equal roles}} \\ 
Hand waving: & \multicolumn{1}{p{13cm}|}{Both: Lifting the right arm into an upright, ninety-degree angle with the open palm facing the partner; moving the lower arm  sideways in an oscillatory motion (around 3-6 cycles); lowering the arm. } \\   
Hand Shaking: & \multicolumn{1}{p{13cm}|}{Both: Stretching the right arm forward to meet the partner's hand, grasping the partners hand; moving the lower arm  up and down in an oscillatory motion (around 3-6 cycles); releasing the partner's hand, lowering the arm.  } \\ 
\multicolumn{2}{l}{\textbf{Leader-follower roles}} \\
Parachute fist-bump: & \multicolumn{1}{p{13cm}|}{Both: Stretching the right arm upwards with the hand closed to a fist to meet the partner's hand, touching the partner's fist with one's own; \newline
Leader: (parachute) Opening the hand and tilting it so that the flat, inner palm faces downwards; keeping the hand above the follower's hand; moving the hand in a slight sideways oscillatory motion while simultaneously moving downwards; \newline
Follower (person): Keeping the hand closed and slightly below the leader's hand; following the slight sideways oscillatory motion of the leader and moving the hand downwards; \newline
    Both: Lowering the arm when the hand is approximately on the height of the hip.} \\   
Rocket fist-bump: & \multicolumn{1}{p{13cm}|}{Both: Stretching the right arm downwards with the hand closed to a fist to meet the partner's hand, touching the partner's fist with one's own; \newline
Leader (rocket): Opening the hand slightly to point to fingers upwards; keeping the hand above the follower's hand; moving the hand upwards; \newline
Follower (fire): Opening the hand with all fingers stretched downwards;  keeping the hand below the leader's hand; wiggling the fingers to simulate fire; moving the hand upwards; \newline
Both: Lowering the arm when the hand is approximately on the height of the shoulders.} 
\end{tabular}}
\end{table}

Between actions, the two partners are standing in an upright position with both arms directed downwards close to the body. 

As the robot is not necessarily equipped with a hand-like gripper, the actions involving finger movement are omitted during human-robot interaction. Furthermore, we assume the robot to take the role of the follower.

\begin{figure}[t]
    \centering
    \begin{minipage}[t]{0.465\textwidth}
        \centering
                \includegraphics[width=\textwidth]{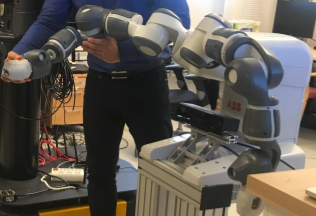}
                a)
    \end{minipage}
    \begin{minipage}[t]{0.4\textwidth}
        \centering
                \includegraphics[width=\textwidth]{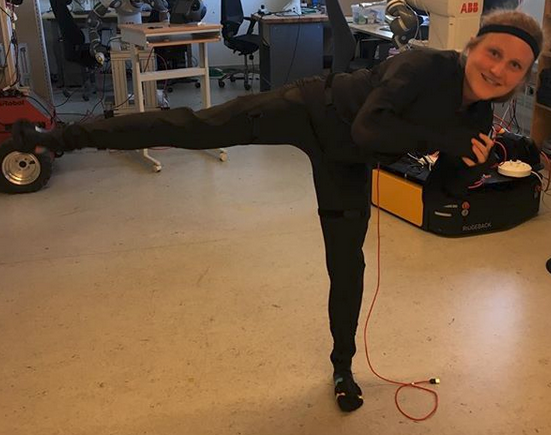} 
                b)
    \end{minipage}
    \caption{\textbf{a)} The right arm of the Yumi robot used in the experiments. \textbf{b)} A rokoko smart suit in action.  }
    \label{fig:data_collection}
\end{figure}

\subsection{Data collection}

We collected data from human-human and human-robot interaction respectively. The robotic setup and the human motion recording setup are described below, followed by the data collection procedure.

\subsubsection{Robotic system setup}

In this work, we use a dual-armed YuMi-IRB 14000 robot which has been developed by ABB specifically with human-robot collaboration in mind. As depicted in Figure \ref{fig:data_collection} a), each arm has seven joints Arm 1 (rotation motion), Arm 2 (bend motion), Arm 7 (rotation motion), Arm 3 (bend motion), Wrist 4 (rotation motion), Wrist 5 (bend motion) and Flange 6 (rotation motion). To control the robot, we work in the joint angle space, i.e. at each time step we have access to a seven dimensional vector consisting of radial measurements. To control the robot, we provide the system with the next expected joint angle configuration or a whole trajectory thereof. We sample the robot's joint angles at a frequency of 40 Hz. 

\begin{table}[b]
\centering
\begin{tabular}{ |c|c|c|c|c|c|c| } 
 \hline
   & \multicolumn{3}{c}{Human-human data} &  \multicolumn{3}{|c|}{Human-robot data} \\
   \hline
 Action type  & \# trials & min. duration & max duration & \# trials & min. duration & max duration \\
 \hline
 hand shake & 38 & 8.5 s & 12.5s & 10 & 10.4 s & 14.5 s\\
 hand wave & 31 & 8.5 s & 17.5 s & 10 & 12.7 s & 17.4 s\\
 parachute & 49 & 7.0 s & 12.0 s & 11 & 11.0 s & 14.3 s\\
 rocket & 70 & 3.0 s & 6.0 s     & 10 & 11.1 s & 13.8 s\\
 \hline 
\end{tabular}
\caption{Statistics of the collected dataset.}
\label{tab:data_statistics}
\end{table}

\subsubsection{Human motion capture}

We recorded the 3D position of the human joints in Cartesian space  during interaction with help of two Rokoko smart suits\footnote{\url{https://www.rokoko.com/}}. As shown in Figure \ref{fig:data_collection} b), these textile suits are equipped with 19 inertia sensors with which motion is recorded. Via wireless communication with a Wi-fi access point, the suits are able to record whole-body movements at a rate of up to 100 Hz. While simultaneous recordings with several suits are possible, we  align the recordings offline. We record the 3D Cartesian positions of each joint in meters with respect to a body-centric reference frame. The data is sampled down to match the 40 Hz of the robot recording.

\subsubsection{Collection procedure}
 
\begin{figure}[t]
    \centering
    \includegraphics[width=\textwidth]{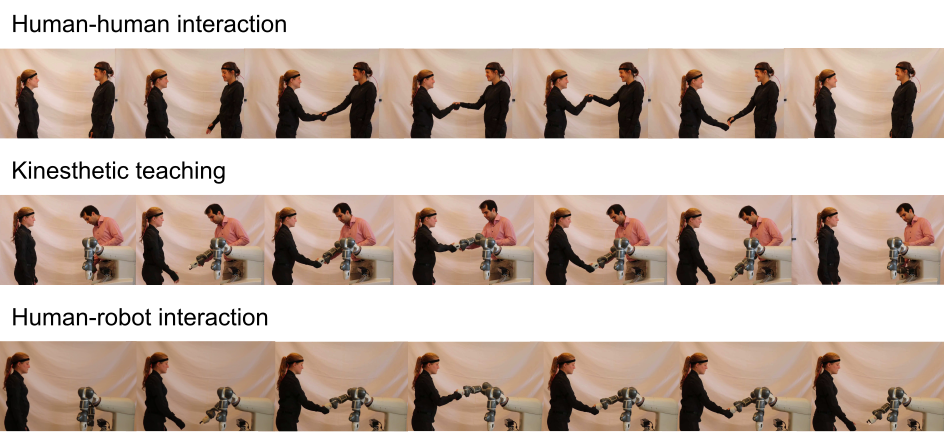}
    \caption{The data collected during human-human interaction (top) and kinesthetic teaching (middle) is used train the proposed models. These are employed in human-robot interaction tasks (bottom). }
    \label{fig:allimages}
\end{figure}

For the human-human dataset, we asked two participants to perform all four actions as described in Section \ref{sec:task_description} for approximately six minutes each. The exact number of repetitions of each action type as well as duration statistics are listed in Table \ref{tab:data_statistics}. A recording of the action \textit{hand-shake} is depicted in the top of Figure \ref{fig:allimages}.

For the robot-human dataset, we asked one of the participants to perform all for actions together with the robot. To this end, we made use of kinesthetic teaching, i.e. a human expert guided the arm of the robot during the interaction. As shown in Table \ref{tab:data_statistics}, the duration of the human-robot trials is on average slightly longer than the human-human trials. A recording of the action \textit{hand-shake} is depicted in the middle of Figure \ref{fig:allimages}.

% ############################################################################################
% --------------------------------------------------------------------------------------------
% Results
% --------------------------------------------------------------------------------------------
% ############################################################################################
\section{Results}

In this section we present the performance of the proposed approach. 
Online employment of our approach during the action \textit{hand-shake} is depicted in the bottom of Figure \ref{fig:allimages}. More examples can be found here youtube.
In the analysis we present results on held-out test datasets. We begin by investigating the predictive performance of the models trained on the human-human dataset. This will be followed by an analysis of the robot motion prediction. In this case, we consider both the predictive error as well as the entrainment of predicted vs ground truth robot motion with the human motion.

\subsection{Predictive performance on human-human data}
\label{sec:predicitive_hh}

\begin{figure}[t]
    \centering
    \includegraphics[width=\textwidth]{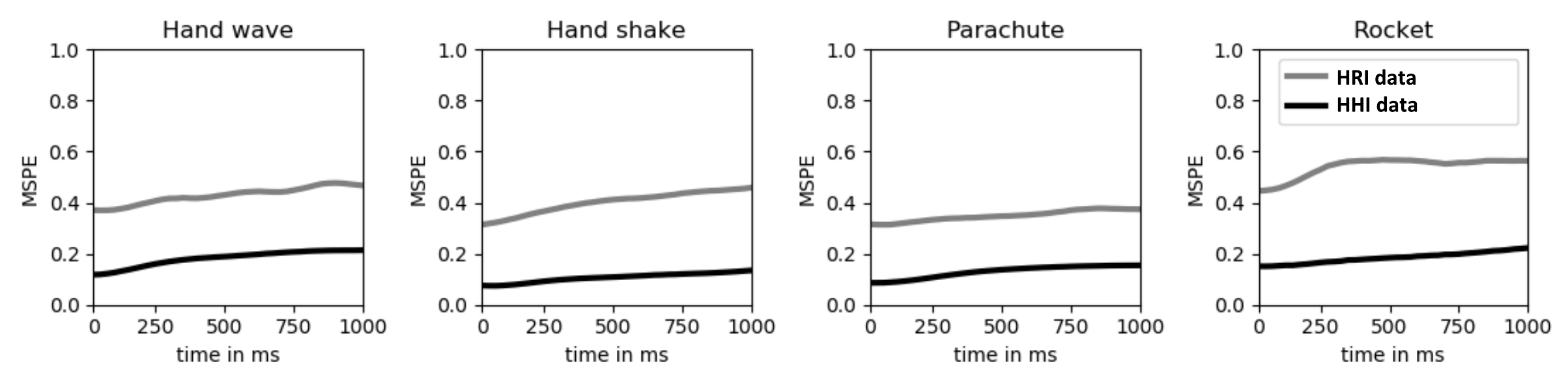}
    \caption{The mean squared prediction error (MSPE) in meters for human-human interaction over a time horizon of one second.}
    \label{fig:results_mse_hh}
\end{figure}

We have two reasons for collecting additional human-human interaction data. Firstly, we hypothesize that the dynamics learned based on HHI data can guide robot action selection during HRI experiments. Secondly, it is easier to collect HHI data, allowing for larger datasets. To test the second hypothesis we trained the human motion embedding and dynamics models both on HHI data and only on the human data contained in the HRI data. In the latter case, the dynamics variable is not restricted to match a human partner. We test the predictive capacity of both these models by computing the mean squared prediction error (MSPE) for the time window $w$ on both test data sets (HRI and HHI). The results are depicted in Figure \ref{fig:results_mse_hh}. Two observations can be made. First of all, the model trained on HRI data does not generalize well, mainly caused by the small training data set. Secondly, the prediction error does not increase drastically over time as should be expected. Due to the fact that we do not force the model to predict a whole trajectory as e.g. \cite{butepage2018anticipating} but only a latent variable which can be decoded into a trajectory, our model is less prone to regress to the mean but to encode the actual motion.

\subsection{Predictive performance on human-robot data}
\label{sec:predicitive_hr}

In this section we inspect how our proposed dynamics transfer approach performs against the baselines. As the different joints move to different extents, the range of joint angles varies. Therefore we measure the predictive error not with the MSPE as in the case of HHI predictions but with the normalized root-mean-square deviation (NRMSD) which is computed as follows:
\begin{equation}
NRMSD(\{\mathbf{x}^r_{1:T_{tr}, j}\}_{tr \in 1:TR}, \{\mathbf{\hat{x}}^r_{1:T_{tr}, j}\}_{tr \in 1:TR}) = \frac{1}{TR} \sum_{tr \in 1:TR} \sqrt{\frac{1}{T_{tr} (j_{max} - j_{min})} \sum_{t=1}^{T_{tr}} (\mathbf{x}^r_{t,j} - \mathbf{\hat{x}}^r_{t,j})^2},
\end{equation}
for the $jth$ joint. Here $tr$ denotes trial, i.e. one execution of an interaction, and $TR$ is the number of trials. $j_{max} $ and $ j_{min}$ denote the maximum and minimum value that has been recorded for the $jth$ joint in the training data.

\begin{figure}[t]
    \centering
        \includegraphics[width=\textwidth]{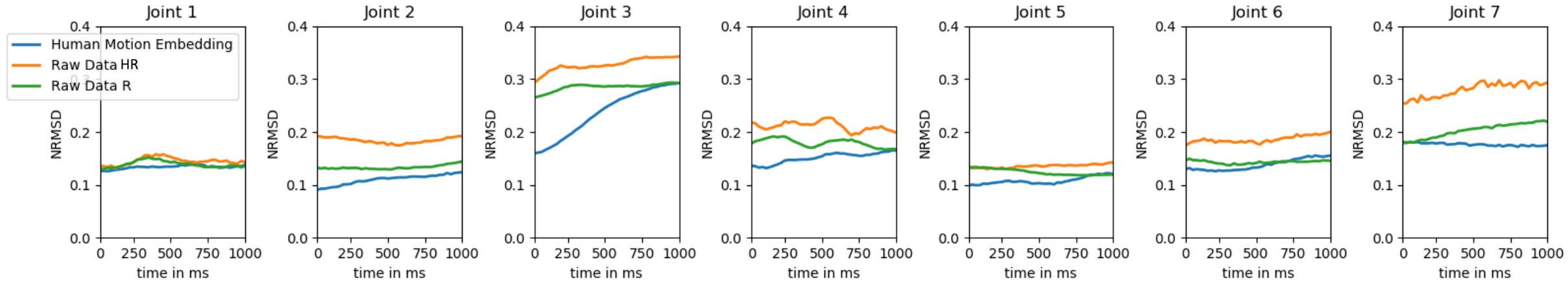}
    \caption{The normalized root-mean-square deviation (NRMSD) for robot motion during human-robot interaction over a time horizon of one second.}
    \label{fig:results_hr_error}
\end{figure}

\begin{table}[b]
\centering
\begin{tabular}{ |c|c|c|c| } 
 \hline
     \multicolumn{4}{|c|}{NRMSD computed on robot testing data}   \\
   \hline
 Human Motion Embedding & Raw Data RH & Raw Data R & Gaussian model \\
 \hline
   \textbf{0.16}  & 0.22 & 0.18  & 0.20 \\
 \hline 
\end{tabular}
\caption{NRMSD computed on robot testing data averaged over all joints.}
\label{tab:robot_comp}
\end{table}

\begin{figure}[t]
    \centering
    \includegraphics[width=\textwidth]{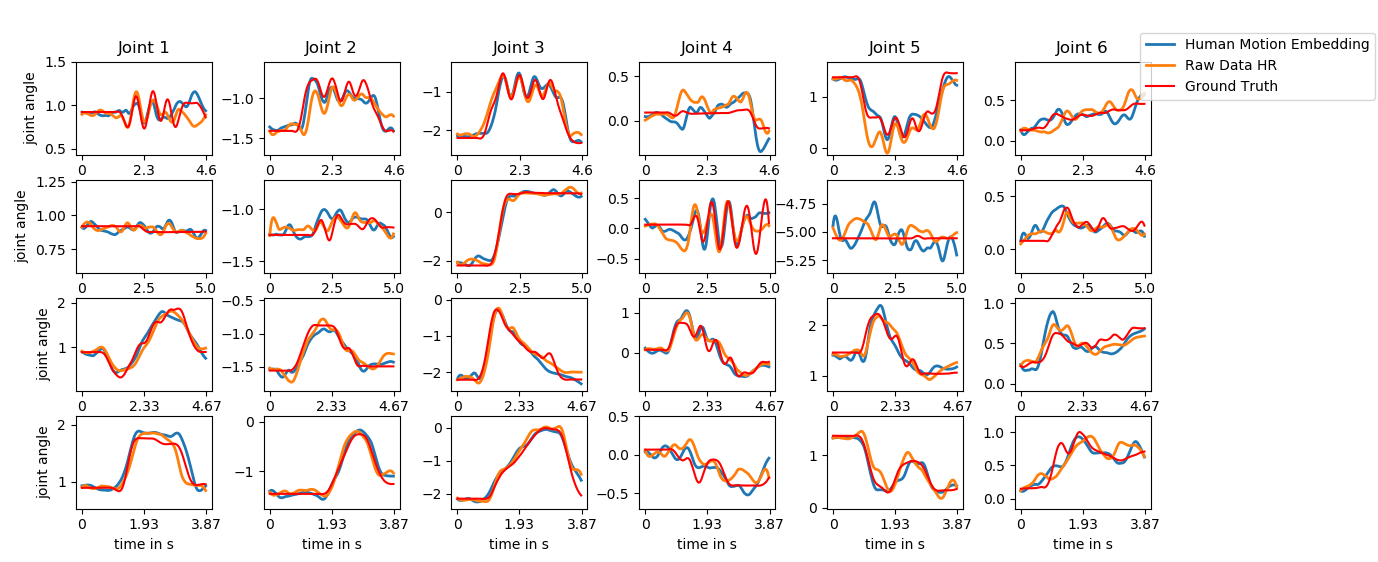}
    \caption{The joint angle trajectory of joint 1 to 6 for a testing trial of each of the actions \textit{hand-wave, hand-shake, parachute} and \textit{rocket} (top to bottom).}
    \label{fig:predictions}
\end{figure}

We start by comparing our approach (Human Motion Embedding) to the two models that have an identical structure but that differ in the type of input data (Raw Data HR and Raw Data R). To this end, we provide ten time steps as input to the models and let the recurrent network predict thirty steps as described in Section \ref{sec:motion_generation}. This process is repeated until the end of a trial is reached.  Since the Raw Data HR model is not able to generate human motion, we provide it with the last observed human pose.  Through the motion embedding, the models produce a prediction of the next forty time steps (1 second). We average over all time steps and present the results in Figure \ref{fig:results_hr_error}. The Human Motion Embedding appears to produce the smallest errors, especially for those joints that are vital for the interaction (joint 2, 3 and 4). The wrist joints (joint 6 and 7) are of less importance and do also show a larger degree of between-trial variance in the training data. We depict the predictions of each of the Human Motion Embedding model, the Raw Data HR model and the ground truth trajectory for one testing trial of each action in Figure \ref{fig:predictions}.

When averaged over the forty time steps of prediction, the difference becomes clear in Table \ref{tab:robot_comp}, where we also include the Gaussian model. As the Raw Data HR model is not able to predict human motion, it produced the largest error. The Human Motion Embedding outperforms both the adaptive   Raw Data HR and Raw Data R models as well as the non-adaptive Gaussian model. The adaptive Raw Data R model produces a smaller error than the non-adaptive Gaussian model, which also is trained on raw robot data. We will investigate the difference  between adaptive and non-adaptive approaches in more detail in the next section.

\begin{figure}[b]
    \centering
    \includegraphics[width=0.7\textwidth]{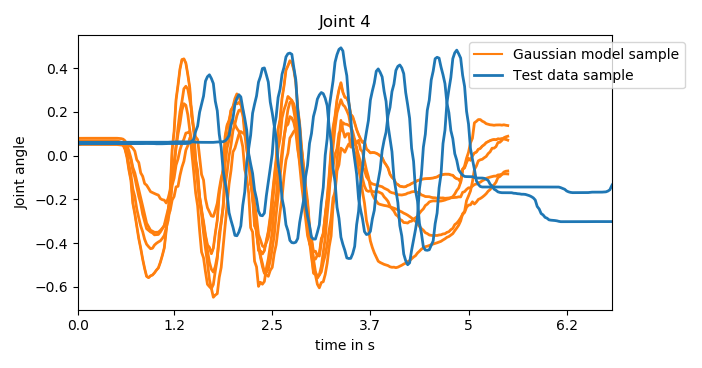}
    \caption{Five samples from the Gaussian model of the hand-shake action and two testing data trials.}
    \label{fig:shaking}
\end{figure} 

\subsection{Non-adaptive vs adaptive motion generation}
\label{sec:non_adaptive}

As discussed in Section \ref{sec:requirementts}, Human-Robot interaction has additional requirements compared to traditional imitation learning. It does not suffice to  learn  a distribution over the trajectories observed in the training data and sample a whole trajectory during run-time. Instead, HRI requires adaptive and predictive models that react to the human's actions such that a sensorimotor coupling between human and robot can arise. We visualize this in Figure \ref{fig:shaking} by sampling from the Gaussian model of joint 4 for the action hand-shake. It becomes apparent that none of the samples is in accordance with any of the testing trials that are also depicted. First of all, the motion onset differs and the duration of the trajectory is predetermined due to the time alignment, while the duration of natural interaction differs from trial to trial. Additionally, the movement is not adapted to the human's hand-shake but has different degrees of phase shift. If we compare these predictions to the predictions of joint 4 in the second row of Figure \ref{fig:predictions}, we realize that the adaptive approach reacts in a timely manner and follows the oscillations of the the ground truth motion that match the human motion. We will investigate the degree of entrainment of the predictions of robot with the human motion in the next section.

\subsection{Entrainement on human-robot data}
\label{sec:entrainment}

\begin{figure}[t]
    \centering
    \includegraphics[width=0.6\textwidth]{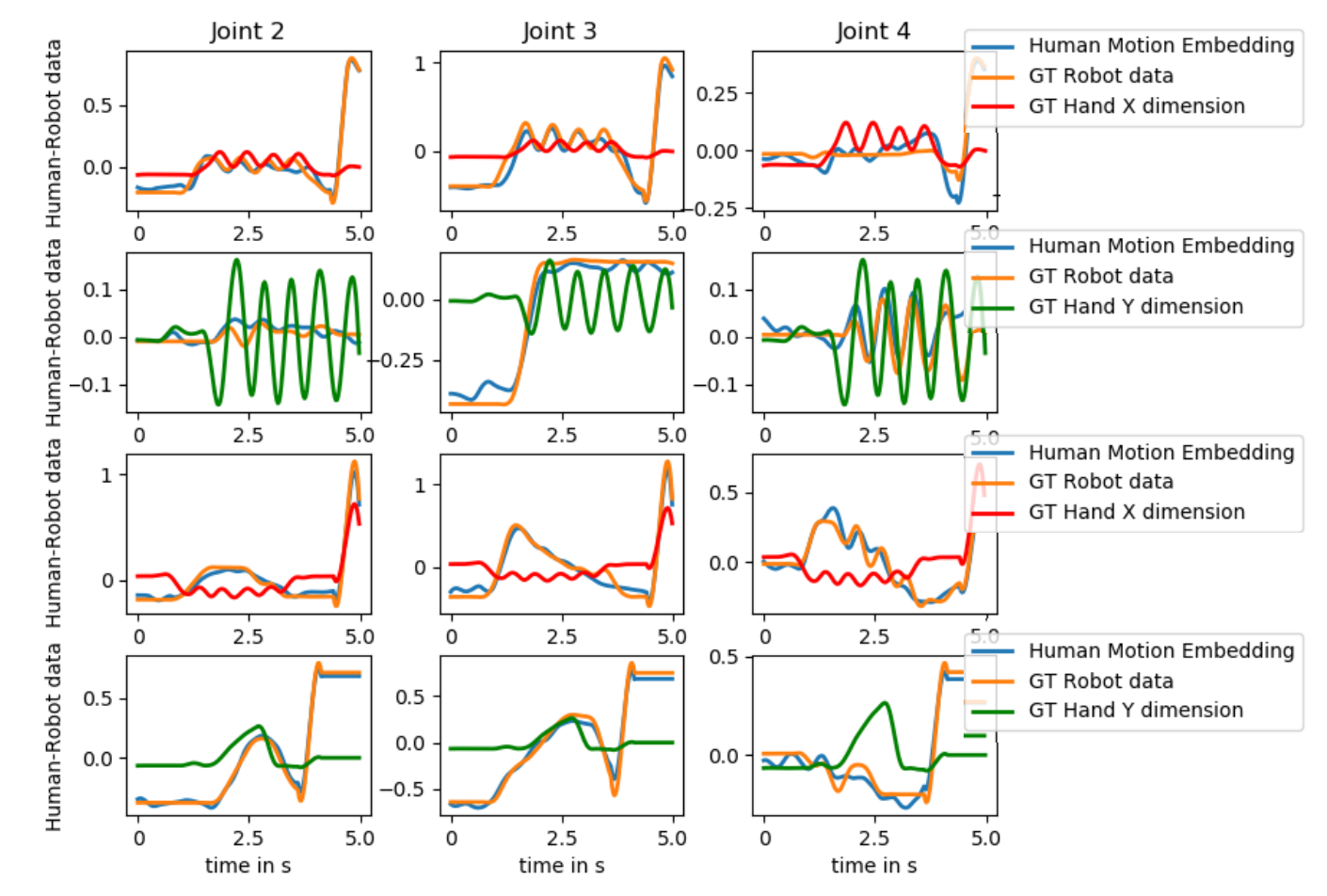}
    \caption{The predictions of the Human Motion Embedding model as well as the ground truth robot motion data and the hand position (X or Y dimension) of the human for a testing trial of each interaction \textit{hand-wave, hand-shake, parachute} and \textit{rocket} (top to bottom). The values are normalized to facilitate comparison.}
    \label{fig:hri}
\end{figure}

With this work we are aiming at developing models that allow for sensorimotor coupling between humans and robots to benefit physical HRI. We visualize the generated predictions of the Human Motion Embedding model as well as the ground truth robot motion data and the hand position of the human for a testing trial of each interaction in Figure \ref{fig:hri}. As not all joints are relevant to a task, we visualize joint 2-4. We see that the predicted motion follows the oscillatory movement of the human hand during hand-wave (see joint 3), hand-shake (see joint 4) and the parachute (see joint 4)  interaction as well the rise and fall of the rocket action (see joint 2 and 3).

\begin{figure}[t]
    \centering
    \includegraphics[width=\textwidth]{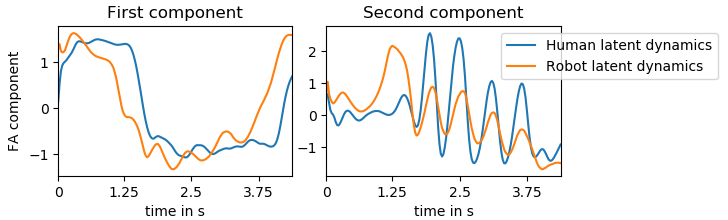}
    \caption{The first two factor analysis (FA) components of a testing trial of the hand-shake interaction computed on both the latent variables extracted from the human ground truth motion and the latent variables predicting the robot motion.}
    \label{fig:latent}
\end{figure}

To investigate whether the models capture this coupling, we extract the dynamics variables of the human motion of an entire testing trial of the hand-shake interaction as well as the latent variables that predict the robot motion. We then apply factor analysis to these two streams of data and compare the two first components to each other. The two components are visualized in Figure \ref{fig:latent}. The first factor appears to represent the general onset, duration and offset of the interaction while the second factor encodes the oscillatory motion of the hand and arm. We see that, although the factor analysis is performed independently on the human and robot latent variables, the overall structure is similar. Additionally, the oscillatory motion is overlapping, indicating a coupling between the two systems.

\section{Conclusion}

In this work, we propose a deep generative model approach to imitation learning of interactive tasks. Our contribution is a novel probabilistic latent variable model which does not predict in joint space but in latent space, which minimizes the chance of regression to the mean. We employ this model both as a dynamics extractor of HHI as well as the basis for the motion generation of a robotic partner. Our experiments indicate that HRI requires adaptive models which take the human motion and task dynamics into account. These dynamics, which encode the movement of both humans, see Figure \ref{fig:dynamics}, and therefore the coupling of the human partners during interaction, guide the generation of the robot which thus is coupled to its human partner.

After having established that the cheaper HHI data is required for high predictive performance, see Section \ref{sec:predicitive_hh}, we demonstrate that the extracted dynamics facilitate the performance of the predictive model of robot motion, see Section \ref{sec:predicitive_hr}. This indicates that the encoding of the future human motion and task dynamics can contribute to the robot's motion planning. This is in contrast to common approaches to imitation learning for interaction which use non-adaptive models. As we discuss in Section \ref{sec:non_adaptive}, a non-adaptive trajectory model does not suffice in interactive tasks such as hand-shaking. With help of our generative approach, we can create synchronized behavior which shows a level of entrainment between human and robot, see Section \ref{sec:entrainment}.  

We believe that prediction and adaptation are essential to allow for natural HRI in shared workspaces. In future work, we plan to employ the system in real-time and to extend it to more complex tasks.

\section*{Author Contributions}

Judith B\"utepage contributed to the idea development and data collection, developed the methodology, implemented and trained the models, evaluated the models and wrote the manuscript.
Ali Ghadirzadeh contributed to the idea, data collection and development of the robot software. 
\"Ozge \"Oztimur Karada\~g contributed to the data collection and implementation of other baselines.
M\aa rten Bj\"orkman and
Danica Kragic supervised the work. 

\section*{Acknowledgments}
This work was supported by the EU through the project socSMCs (H2020-FETPROACT-2014) and the Swedish Foundation for Strategic Research.

\bibliographystyle{IEEEtran}
\bibliography{frontier}

\end{document}